%% file: main.tex
\documentclass[runningheads]{llncs}
\usepackage{graphicx}
\usepackage{amsmath,amssymb} 
\usepackage{color}
\usepackage[width=122mm,left=12mm,paperwidth=146mm,height=193mm,top=12mm,paperheight=217mm]{geometry}

\usepackage{mathtools}
\usepackage{booktabs}
\usepackage[caption=false]{subfig}

\usepackage{import}
\usepackage{minibox}

\usepackage{mynotation}

\usepackage[pagebackref=true,breaklinks=true,letterpaper=true,colorlinks,bookmarks=false]{hyperref}

\newcommand{\parsection}[1]{\noindent\textbf{#1:}}

\begin{document}
\pagestyle{headings}
\mainmatter

\title{Beyond Correlation Filters: Learning Continuous Convolution Operators for Visual Tracking} 

\titlerunning{Learning Continuous Convolution Operators for Visual Tracking}

\authorrunning{Danelljan et al.}

\author{Martin Danelljan, Andreas Robinson, Fahad Shahbaz Khan, Michael Felsberg}
\institute{CVL, Department of Electrical Engineering, Link\"oping University, Sweden\\
	\email{\{martin.danelljan, andreas.robinson, fahad.khan, michael.felsberg\}@liu.se}}

\maketitle

\begin{abstract}
	\input{abstract}
\end{abstract}

\input{introduction}

\input{relatedwork}

\input{method}

\input{experiments}

\input{conclusions}

\bibliographystyle{splncs03}
\bibliography{references}
\end{document}

%% file: abstract.tex
Discriminative Correlation Filters (DCF) have demonstrated excellent performance for visual object tracking. The key to their success is the ability to efficiently exploit available negative data by including all shifted versions of a training sample. However, the underlying DCF formulation is restricted to single-resolution feature maps, significantly limiting its potential. 
In this paper, we go beyond the conventional DCF framework and introduce a novel formulation for training \emph{continuous} convolution filters. We employ an implicit interpolation model to pose the learning problem in the continuous spatial domain.
Our proposed formulation enables efficient integration of multi-resolution deep feature maps, leading to superior results on three object tracking benchmarks: OTB-2015 ($+5.1\%$ in mean OP), Temple-Color ($+4.6\%$ in mean OP), and VOT2015 ($20\%$ relative reduction in failure rate). Additionally, our approach is capable of sub-pixel localization, crucial for the task of accurate feature point tracking. We also demonstrate the effectiveness of our learning formulation in extensive feature point tracking experiments. Code and supplementary material are available at \url{http://www.cvl.isy.liu.se/research/objrec/visualtracking/conttrack/index.html}.

%% file: introduction.tex
\section{Introduction}

Visual tracking is the task of estimating the trajectory of a target in a video. It is one of the fundamental problems in computer vision. Tracking of objects or feature points has numerous applications in robotics, structure-from-motion, and visual surveillance. In recent years, Discriminative Correlation Filter (DCF) based approaches have shown outstanding results on object tracking benchmarks \cite{VOT2014,OTB2015}. DCF methods train a correlation filter for the task of predicting the target classification scores. Unlike other methods, the DCF efficiently utilize all spatial shifts of the training samples by exploiting the discrete Fourier transform.

Deep convolutional neural networks (CNNs) have shown impressive performance for many tasks, and are therefore of interest for DCF-based tracking.
A CNN consists of several layers of convolution, normalization and pooling operations.
Recently, activations from the last convolutional layers have been successfully employed for image classification. 
Features from these deep convolutional layers are discriminative while preserving spatial and structural information. Surprisingly, in the context of tracking, recent DCF-based methods \cite{DanelljanVOT2015,HCF_ICCV15} have demonstrated the importance of shallow convolutional layers. These layers provide higher spatial resolution, which is crucial for accurate target localization. However, fusing multiple layers in a DCF framework is still an open problem.

\begin{figure}[!t]
	\centering%
	\vspace{-8mm}
	\newcommand{\wid}{0.93\textwidth}%
	\def\svgwidth{\wid}%
	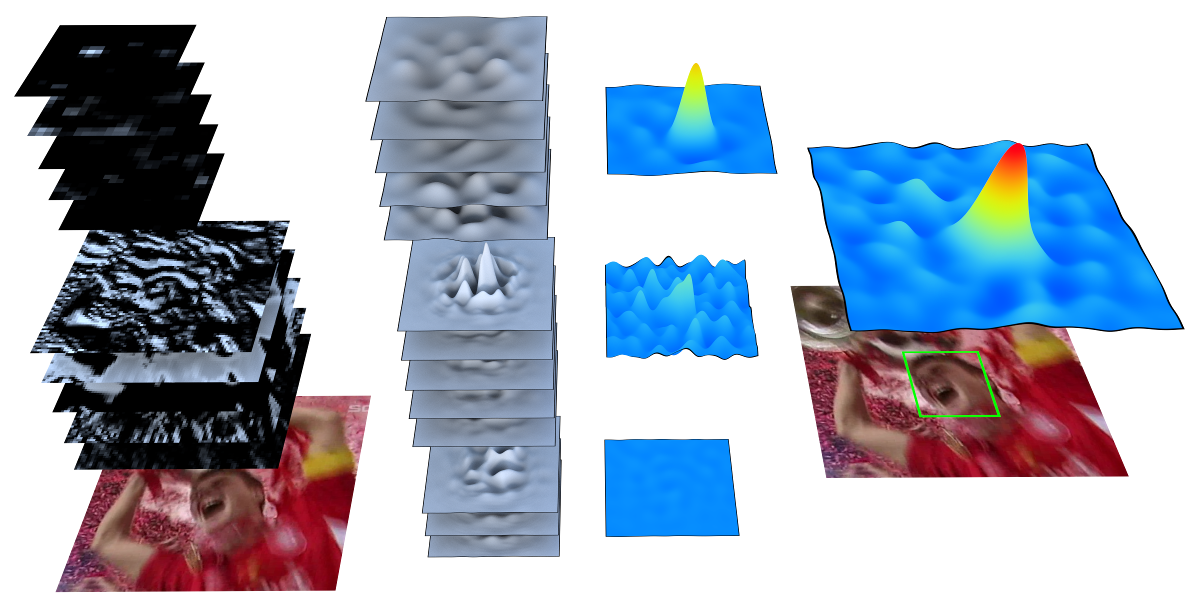%
	\vspace{2.0mm}%
	\caption{Visualization of our continuous convolution operator, applied to a multi-resolution deep feature map. The feature map (\textit{left}) consists of the input RGB patch along with the first and last convolutional layer of a pre-trained deep network. The second column visualizes the continuous convolution filters learned by our framework. The resulting continuous convolution outputs for each layer (third column) are combined into the final continuous confidence function (\textit{right}) of the target (green box).}\vspace{-3.5mm}
	\label{fig:intro}
\end{figure}

The conventional DCF formulation is limited to a single-resolution feature map. Therefore, all feature channels must have the same spatial resolution, as in e.g.\ the HOG descriptor. This limitation prohibits joint fusion of multiple convolutional layers with different spatial resolutions.
A straightforward strategy to counter this restriction is to explicitly resample all feature channels to the same common resolution. However, such a resampling strategy is both cumbersome, adds redundant data and introduces artifacts.
Instead, a principled approach for integrating multi-resolution feature maps in the learning formulation is preferred.

In this work, we propose a novel formulation for learning a convolution operator in the \emph{continuous} spatial domain. The proposed learning formulation employs an implicit interpolation model of the training samples. Our approach learns a set of convolution filters to produce a \emph{continuous-domain} confidence map of the target. This enables an elegant fusion of multi-resolution feature maps in a joint learning formulation. Figure~\ref{fig:intro} shows a visualization of our continuous convolution operator, when integrating multi-resolution deep feature maps. 
We validate the effectiveness of our approach on three object tracking benchmarks: OTB-2015 \cite{OTB2015}, Temple-Color \cite{TempleColor} and VOT2015 \cite{VOT2015}. On the challenging OTB-2015 with 100 videos, our object tracking framework improves the state-of-the-art from $77.3\%$ to $82.4\%$ in mean overlap precision.

In addition to multi-resolution fusion, our continuous domain learning formulation enables accurate sub-pixel localization. This is achieved by labeling the training samples with sub-pixel precise continuous confidence maps. Our formulation is therefore also suitable for accurate feature point tracking. Further, our learning-based approach is discriminative and does not require explicit interpolation of the image to achieve sub-pixel accuracy. We demonstrate the accuracy and robustness of our approach by performing extensive feature point tracking experiments on the popular MPI Sintel dataset \cite{butler2012naturalistic}.

%% file: intro.pdf_tex
\begingroup%
  \makeatletter%
  \providecommand\color[2][]{%
    \errmessage{(Inkscape) Color is used for the text in Inkscape, but the package 'color.sty' is not loaded}%
    \renewcommand\color[2][]{}%
  }%
  \providecommand\transparent[1]{%
    \errmessage{(Inkscape) Transparency is used (non-zero) for the text in Inkscape, but the package 'transparent.sty' is not loaded}%
    \renewcommand\transparent[1]{}%
  }%
  \providecommand\rotatebox[2]{#2}%
  \ifx\svgwidth\undefined%
    \setlength{\unitlength}{1200bp}%
    \ifx\svgscale\undefined%
      \relax%
    \else%
      \setlength{\unitlength}{\unitlength * \real{\svgscale}}%
    \fi%
  \else%
    \setlength{\unitlength}{\svgwidth}%
  \fi%
  \global\let\svgwidth\undefined%
  \global\let\svgscale\undefined%
  \makeatother%
  \begin{picture}(1,0.54739372)%
    \put(0,0){\includegraphics[width=\unitlength,page=1]{intro.png}}
    \scriptsize
    \newcommand{\hgt}{-0.026}
    {\fontfamily{bch}\selectfont
    \put(0.05,\hgt){\minibox[c]{Multi-resolution deep \\ feature map}}
    \put(0.30,\hgt){\minibox[c]{Learned continuous\\ convolution filters}}
    \put(0.51,\hgt){\minibox[c]{Confidence scores\\ for each layer}}
    \put(0.72,\hgt){\minibox[c]{Final continuous confidence\\ output function}}
	}
  \end{picture}%
\endgroup%

%% file: relatedwork.tex
\section{Related Work}

Discriminative Correlation Filters (DCF) \cite{MOSSE2010,DanelljanICCV2015,Henriques14} have shown promising results for object tracking. These methods exploit the properties of circular correlation for training a regressor in a sliding-window fashion. Initially, the DCF approaches \cite{MOSSE2010,Henriques12} were restricted to a single feature channel. The DCF framework was later extended to multi-channel feature maps \cite{BoddetiCVPR13,DanelljanCVPR14,galoogahiICCV13}. The multi-channel DCF allows high-dimensional features, such as HOG and Color Names, to be incorporated for improved tracking. In addition to the incorporation of multi-channel features, the DCF framework has been significantly improved lately by, e.g., including scale estimation \cite{DanelljanBMVC14,Li2014}, non-linear kernels \cite{Henriques12,Henriques14}, a long-term memory \cite{LTC_CVPR15}, and by alleviating the periodic effects of circular convolution \cite{DanelljanICCV2015,BoddetiPAMI2015,GaloogahiCVPR2015}. 

With the advent of deep CNNs, fully connected layers of the network have been commonly employed for image representation \cite{OquabCVPR2014,SimonyanICLR2015}. Recently, the last (deep) convolutional layers were shown to be more beneficial for image classification \cite{VedaldiCVPR2015,HengelCVPR2015}. On the other hand, the first (shallow) convolutional layer was shown to be more suitable for visual tracking, compared to the deeper layers \cite{DanelljanVOT2015}. The deep convolutional layers are discriminative and possess high-level visual information. In contrast, the shallow layers contain low-level features at high spatial resolution, beneficial for localization. Ma et al.\ \cite{HCF_ICCV15} employed multiple convolutional layers in a hierarchical ensemble of independent DCF trackers. Instead, we propose a novel continuous formulation to fuse multiple convolutional layers with different spatial resolutions in a \emph{joint} learning framework. 

Unlike object tracking, feature point tracking is the task of accurately estimating the motion of distinctive key-points. It is a core component in many vision systems \cite{Badino2013,klein07parallel,OrvenICRA2015,ZografosACCV2014}. Most feature point tracking methods are derived from the classic Kanade-Lucas-Tomasi (KLT) tracker \cite{LukasKanade81,tomasi1991KLTtracker}. The KLT tracker is a generative method, that is based on minimizing the squared sum of differences between two image patches. In the last decades, significant effort has been spent on improving the KLT tracker \cite{BakerIJCV2004,Fusiello1999}. In contrast, we propose a discriminative learning based approach for feature point tracking.

\parsection{Our approach}
Our main contribution is a theoretical framework for learning discriminative convolution operators in the continuous spatial domain. Our formulation has two major advantages compared to the conventional DCF framework. Firstly, it allows a natural integration of multi-resolution feature maps, e.g.\ combinations of convolutional layers or multi-resolution HOG and color features. This property is especially desirable for object tracking, detection and action recognition applications. Secondly, our continuous formulation enables accurate sub-pixel localization, crucial in many feature point tracking problems.

%% file: method.tex
\section{Learning Continuous Convolution Operators}
\label{sec:method}

In this section, we present a theoretical framework for learning continuous convolution operators. Our formulation is generic and can be applied for supervised learning tasks, such as visual tracking and detection. 

\subsection{Preliminaries and Notation}
\label{sec:preliminaries}

In this paper, we utilize basic concepts and results in continuous Fourier analysis. For clarity, we first formulate our learning method for data defined in a one-dimensional domain, i.e.\ for functions of a single spatial variable. We then describe the generalization to higher dimensions, including images, in section~\ref{sec:higher_dimensions}.

We consider the space $\Lp$ of complex-valued functions $g : \reals \rightarrow \complexes$ that are periodic with period $T > 0$ and square Lebesgue integrable.
The space $\Lp$ is a Hilbert space equipped with an inner product $\langle \cdot , \cdot \rangle$. For functions $g,h \in \Lp$,
\begin{equation}
	\label{eq:definitions}
	\langle g , h \rangle = \frac{1}{T} \integlim{0}{T} g(t) \overline{h(t)} \diff t \, , \qquad g \conv h (t) = \frac{1}{T} \integlim{0}{T} g(t - s) h(s) \diff s \, .
\end{equation}
Here, the bar denotes complex conjugation.
In \eqref{eq:definitions} we have also defined the circular convolution operation $\conv : \Lp \times \Lp \rightarrow \Lp$.

In our derivations, we use the complex exponential functions $e_k(t) = e^{i \frac{2 \pi}{T} k t}$ since they are eigenfunctions of the convolution operation \eqref{eq:definitions}. The set $\{e_k\}_{-\infty}^\infty$ further forms an orthonormal basis for $\Lp$. We define the Fourier coefficients of $g \in \Lp$ as $\hat{g}[k] = \langle g , e_k \rangle$. For clarity, we use square brackets for functions with discrete domains. Any $g \in \Lp$ can be expressed in terms of its Fourier series $g = \sum_{-\infty}^{\infty} \hat{g}[k] e_k$. The Fourier coefficients satisfy Parseval's formula $\|g\|^2 = \| \hat{g} \|^2_{\ell^2}$, where $\|g\|^2 = \langle g , g \rangle$ and $\| \hat{g} \|^2_{\ell^2} = \sum_{-\infty}^{\infty} |\hat{g}[k]|^2$ is the squared $\ell^2$-norm. Further, the Fourier coefficients satisfy the two convolution properties $\widehat{g \conv h} = \hat{g} \hat{h}$ and $\widehat{gh} = \hat{g} \conv \hat{h}$, where $\hat{g} \conv \hat{h} [k] \defeq \sum_{l=-\infty}^{\infty} \hat{g}[k-l] \hat{h}[l]$.

\subsection{Our Continuous Learning Formulation}
\label{sec:learning_formulation}
Here we formulate our novel learning approach. The aim is to train a continuous convolution operator based on training samples $x_j$. The samples consist of feature maps extracted from image patches. Each sample $x_j$ contains $D$ feature channels $x_j^1, \ldots, x_j^D$, extracted from the same image patch. Conventional DCF formulations \cite{DanelljanICCV2015,galoogahiICCV13,Henriques14} assume the feature channels to have the same spatial resolution, i.e.\ have the same number of spatial sample points. Unlike previous works, we eliminate this restriction in our formulation and let $N_d$ denote the number of spatial samples in $x_j^d$. In our formulation, the feature channel $x_j^d \in \reals^{N_d}$ is viewed as a function $x_j^d[n]$ indexed by the discrete spatial variable $n \in \{0, \ldots, N_d-1\}$. The sample space is expressed as $\mathcal{X} = \reals^{N_1} \times \ldots \times \reals^{N_D}$.

To pose the learning problem in the continuous spatial domain, we introduce an implicit interpolation model of the training samples. 
We regard the continuous interval $[0, T) \subset \reals$ to be the spatial support of the feature map. Here, the scalar $T$ represents the size of the support region. In practice, however, $T$ is arbitrary since it represents the scaling of the coordinate system. For each feature channel $d$, we define the interpolation operator $J_d : \reals^{N_d} \rightarrow \Lp$ of the form,
\begin{equation}
	\label{eq:interp_op}
	J_d \big\{x^d\big\} (t) = \sum_{n=0}^{N_d-1} x^d[n] b_d\left(t - \frac{T}{N_d} n\right) .
\end{equation}
The interpolated sample $J_d \big\{x^d\big\} (t)$ is constructed as a superposition of shifted versions of an interpolation function $b_d \in \Lp$. In \eqref{eq:interp_op}, the feature values $x^d[n]$ act as weights for each shifted function.
Similar to the periodic assumption in the conventional discrete DCF formulation, a periodic extension of the feature map is also performed here in \eqref{eq:interp_op}.

As discussed earlier, our objective is to learn a linear convolution operator $S_f : \mathcal{X} \rightarrow \Lp$. This operator maps a sample $x \in \mathcal{X}$ to a target confidence function $s(t) = S_f\{x\}(t)$, defined on the continuous interval $[0,T)$. Here, $s(t) \in \reals$ is the confidence score of the target at the location $t \in [0,T)$ in the image. Similar to other discriminative methods, the target is localized by maximizing the confidence scores in an image region. The key difference in our formulation is that the confidences are defined on a continuous spatial domain. Therefore, our formulation can be used to localize the target with higher accuracy.

In our continuous formulation, the operator $S_f$ is parametrized by a set of convolution filters $f = (f^1, \ldots, f^D) \in \Lp^D$. Here, $f^d \in \Lp$ is the continuous filter for feature channel $d$. We define the convolution operator as,
\begin{equation}
	\label{eq:conv_op}
	S_f\{x\} = \sum_{d=1}^D f^d \conv J_d \big\{x^d\big\} \,,\quad x \in \mathcal{X} \,.
\end{equation}
Here, each feature channel is first interpolated using \eqref{eq:interp_op} and then convolved with its corresponding filter. Note that the convolutions are performed in the continuous domain, as defined in \eqref{eq:definitions}. In the last step, the convolution responses from all filters are summed to produce the final confidence function.

In the standard DCF, each training sample is labeled by a discrete function that represents the desired convolution output. In contrast, our samples $x_j \in \mathcal{X}$ are labeled by confidence functions $y_j \in \Lp$, defined in the continuous spatial domain. Here, $y_j$ is the desired output of the convolution operator $S_f\{x_j\}$ applied to the training sample $x_j$. 
This enables sub-pixel accurate information to be incorporated in the learning. 
The filter $f$ is trained, given a set of $m$ training sample pairs $\{(x_j,y_j)\}_1^m \subset \mathcal{X} \times \Lp$, by minimizing the functional,
\begin{equation}
	\label{eq:loss_spatial}
	E(f) = \sum_{j=1}^{m} \alpha_j \left\| S_f\{x_j\} - y_j \right\|^2 + \sum_{d=1}^{D} \left\| w f^d \right\|^2 \,.
\end{equation}
Here, the weights $\alpha_j \geq 0$ control the impact of each training sample. We additionally include a spatial regularization term, similar to \cite{DanelljanICCV2015}, determined by the penalty function $w$. This regularization enables the filter to be learned on arbitrarily large image regions by controlling the spatial extent of the filter $f$. Spatial regions typically corresponding to background features are assigned a large penalty in $w$, while the target region has small penalty values. Thus, $w$ encodes the prior reliability of features depending on their spatial location. Unlike \cite{DanelljanICCV2015}, the penalty function $w$ is defined on the whole continuous interval $[0,T)$ and periodically extended to $w \in \Lp$. Hence, $\left\| w f^d \right\| < \infty$ is required in \eqref{eq:loss_spatial}. This is implied by our later assumption of $w$ having finitely many non-zero Fourier coefficients $\hat{w}[k]$. Next, we derive the procedure to train the continuous filter $f$, using the proposed formulation \eqref{eq:loss_spatial}.

\subsection{Training the Continuous Filter}
\label{sec:training}
To train the filter $f$, we minimize the functional \eqref{eq:loss_spatial} in the Fourier domain. By using results from Fourier analysis it can be shown\footlabel{supp}{See the supplementary material for a detailed derivation.} that the Fourier coefficients of the interpolated feature map are given by $\widehat{J_d\big\{x^d\big\}}[k] = X^d[k] \hat{b}_d[k]$. Here, $X^d[k] \defeq \sum_{n=0}^{N_d-1} x^d[n] e^{-i \frac{2 \pi}{N_d} n k}, \, k\in \integers$ is the discrete Fourier transform (DFT) of $x^d$. 
By using linearity and the convolution property in section~\ref{sec:preliminaries}, the Fourier coefficients of the output confidence function \eqref{eq:conv_op} are derived as 
\begin{equation}
	\label{eq:score_fs}
	\widehat{S_f\{x\}}[k] = \sum_{d=1}^{D} \hat{f}^d[k] X^d[k] \hat{b}_d[k] \,,\quad k \in \integers \,.
\end{equation}
By applying Parseval's formula to \eqref{eq:loss_spatial} and using \eqref{eq:score_fs}, we obtain
\begin{equation}
	\label{eq:loss_fs}
	E(f) = \sum_{j=1}^{m} \alpha_j \left\| \sum_{d=1}^{D} \hat{f}^d X^d_j \hat{b}_d - \hat{y}_j \right\|^2_{\ell^2} + \sum_{d=1}^{D} \left\|\hat{w} \conv \hat{f^d} \right\|^2_{\ell^2} \,.
\end{equation}
Hence, the functional $E(f)$ can equivalently be minimized with respect to the Fourier coefficients $\hat{f}^d[k]$ for each filter $f^d$. We exploit the Fourier domain formulation \eqref{eq:loss_fs} to minimize the original loss \eqref{eq:loss_spatial}.

For practical purposes, the filter $f$ needs to be represented by a finite set of parameters. One approach is to employ a parametric model to represent an infinite number of coefficients. In this work, we instead obtain a finite representation by minimizing \eqref{eq:loss_fs} over the finite--dimensional subspace $V = \linspan \{e_k\}_{-K_1}^{K_1} \times \ldots \times \linspan \{e_k\}_{-K_D}^{K_D} \subset \Lp^D$. That is, we minimize \eqref{eq:loss_fs} with respect to the coefficients $\{\hat{f}^d[k]\}_{-K_d}^{K_d}$, while assuming $\hat{f}^d[k] = 0$ for $|k| > K_d$. In practice, $K_d$ determines the number of filter coefficients $\hat{f}^d[k]$ to be computed for feature channel $d$ during learning. Increasing $K_d$ leads to a better estimate of the filter $f^d$ at the cost of increased computations and memory consumption. In our experiments, we set $K_d = \left\lfloor \frac{N_d}{2} \right\rfloor$ such that the number of stored filter coefficients for channel $d$ equals the spatial resolution $N_d$ of the training sample $x^d$.

To derive the solution to the minimization problem \eqref{eq:loss_fs} subject to $f \in V$, we introduce the vector of non-zero Fourier coefficients $\vecft{f}^d = (\hat{f^d}[-K_d] \cdots \hat{f^d}[K_d])\tp \in \complexes^{2K_d+1}$ and define the coefficient vector $\vecft{f} = \big[ (\vecft{f}^1)\tp \cdots (\vecft{f}^D)\tp \big]\tp$. Further, we define $\vecft{y}_j = (\hat{y}_j[-K] \cdots \hat{y}_j[K])\tp$ be the vectorization of the $K \defeq \max_d K_d$ first Fourier coefficients of $y_j$. To simplify the regularization term in \eqref{eq:loss_fs}, we let $L$ be the number of non-zero coefficients $\hat{w}[k]$, such that $\hat{w}[k] = 0$ for all $|k| > L$. We further define $W_d$ to be the $(2K_d+2L+1) \times (2K_d+1)$ Toeplitz matrix corresponding to the convolution operator $W_d \vecft{f}^d = \vecop \hat{w} \conv \hat{f}^d$. Finally, let $W$ be the block-diagonal matrix $W = W_1 \oplus \cdots \oplus W_D$. The minimization of the functional \eqref{eq:loss_fs} subject to $f \in V$ is equivalent to the following least squares problem,
\begin{equation}
	\label{eq:loss_ls1}
	E_V(\vecft{f}) = \sum_{j=1}^{m} \alpha_j \left\| A_j \vecft{f} - \vecft{y}_j \right\|^2_2 + \left\| W \vecft{f} \right\|^2_2 \,.
\end{equation}
Here, the matrix $A_j = [A_j^1 \cdots A_j^D]$ has $2K+1$ rows and contains one diagonal block $A_j^d$ per feature channel $d$ with $2K_d+1$ columns containing the elements $\{X^d_j[k] \hat{b}_d[k]\}_{-K_d}^{K_d}$. In \eqref{eq:loss_ls1}, $\|\cdot\|_2$ denotes the standard Euclidian norm in $\complexes^M$.

To obtain a simple expression of the normal equations, we define the sample matrix $A = [A_1\tp \cdots A_m\tp]\tp$, the diagonal weight matrix $\Gamma = \alpha_1 I \oplus \cdots \oplus \alpha_m I$ and the label vector $\vecft{y} = [\vecft{y}_1\tp \cdots \vecft{y}_m\tp]\tp$. The minimizer of \eqref{eq:loss_ls1} is found by solving the normal equations,
\begin{equation}
	\label{eq:normal_eq}
	\left( A\ctp \Gamma A + W\ctp W \right) \vecft{f} = A\ctp \Gamma \vecft{y} \,.
\end{equation}
Here, $\ctp$ denotes the conjugate-transpose of a matrix. Note that \eqref{eq:normal_eq} forms a sparse linear equation system if $w$ has a small number of non-zero Fourier coefficients $\hat{w}[k]$. In our object tracking framework, presented in section~\ref{sec:object_tracking}, we employ the Conjugate Gradient method to iteratively solve \eqref{eq:normal_eq}. For our feature point tracking approach, presented in section~\ref{sec:feature_tracking}, we use a single-channel feature map and a constant penalty function $w$ for improved efficiency. This results in a diagonal system \eqref{eq:normal_eq}, which can be efficiently solved by a direct computation.

\subsection{Desired Confidence and Interpolation Function}
\label{sec:label_and_interp}
Here, we describe the choice of the desired convolution output $y_j$ and the interpolation function $b_d$. We construct both $y_j$ and $b_d$ by periodically repeating functions defined on the real line. In general, the $T$-periodic repetition of a function $g$ is defined as $g_T(t) = \sum_{-\infty}^{\infty} g(t - nT)$. In the derived Fourier domain formulation \eqref{eq:loss_fs}, the functions $y_j$ and $b_d$ are represented by their respective Fourier coefficients. The Fourier coefficients of a periodic repetition $g_T$ can be retrieved from the continuous Fourier transform $\hat{g}(\xi)$ of $g(t)$ as $\hat{g}_T[k] = \frac{1}{T} \hat{g}(\frac{k}{T})$.\footref{supp_det} We use this property to compute the Fourier coefficients of $y_j$ and $b_d$.

To construct the desired convolution output $y_j$, we let $u_j \in [0,T)$ denote the estimated location of the target object or feature point in sample $x_j$. We define $y_j$ as the periodic repetition of the Gaussian function $\exp\big(-\frac{(t-u_j)^2}{2\sigma^2}\big)$ centered at $u_j$. This provides the following expression for the Fourier coefficients,
\begin{equation}
	\label{eq:y_fs}
	\hat{y}_j[k] = \frac{\sqrt{2\pi \sigma^2}}{T} \exp\left(- 2 \sigma^2 \left(\frac{\pi k}{T}\right)^2 - i \frac{2\pi}{T} u_j k\right) \,.
\end{equation}
The variance $\sigma^2$ is set to a small value to obtain a sharp peak. Further, this ensures a negligible spatial aliasing.
In our work, the functions $b_d$ are constructed based on the cubic spline kernel $b(t)$.
The interpolation function $b_d$ is set to the periodic repetition of a scaled and shifted version of the kernel $b\big(\frac{N_d}{T}\big(t - \frac{T}{2N_d}\big)\big)$, to preserve the spatial arrangement of the feature pyramid. The Fourier coefficients of $b_d$ are then obtained as $\hat{b}_d[k] = \frac{1}{N_d} \exp\big(-i \frac{\pi}{N_d} k\big) \hat{b} \big(\frac{k}{N_d}\big)$.\footlabel{supp_det}{Further details are given in the supplementary material.}

\subsection{Generalization to Higher Dimensions}
\label{sec:higher_dimensions}
The proposed formulation can be extended to domains of arbitrary number of dimensions. For our tracking applications we specifically consider the two-dimensional case, but higher-dimensional spaces can be treated similarly. For images, we use the space $L^2(T_1,T_2)$ of square-integrable periodic functions of two variables $g(t_1,t_2)$. 
The complex exponentials are then given by $e_{k_1,k_2}(t_1,t_2) = e^{i \frac{2 \pi}{T_1} k_1 t_1} e^{i \frac{2 \pi}{T_2} k_2 t_2}$. For the desired convolution output $y_j$, we employ a 2-dimensional Gaussian function. Further, the interpolation functions are obtained as a separable combination of the cubic spline kernel, i.e.\ $b(t_1,t_2) = b(t_1)b(t_2)$. The derivations presented in section~\ref{sec:training} also hold for the higher dimensional cases.

\section{Our Tracking Frameworks}
We apply our continuous learning formulation for two problems: visual object tracking and feature point tracking. We first present the localization procedure, which is based on maximizing the continuous confidence function. This is shared for both the object and feature point tracking frameworks.

\subsection{Localization Step}
\label{sec:localization}
Here, the aim is to localize the tracked target or feature point using the learned filter $f$. This is performed by first extracting a feature map $x \in \mathcal{X}$ from the region of interest in an image. The Fourier coefficients of the confidence score function $s = S_f\{x\}$ are then calculated using \eqref{eq:score_fs}. We employ a two-step approach for maximizing the score $s(t)$ on the interval $t \in [0,T)$. To find a rough initial estimate, we first perform a \emph{grid search}, where the score function is evaluated at the discrete locations $s \big( \frac{T n}{2K+1} \big)$ for $n = 0, \ldots, 2K$. This is efficiently implemented as a scaled inverse DFT of the non-zero Fourier coefficients $\hat{s}[k], k = -K, \ldots, K$. The maximizer obtained in the grid search is then used as the initialization for an iterative optimization of the Fourier series expansion $s(t) = \sum_{-K}^{K} \hat{s}[k] e_k(t)$. We employ the standard Newton's method for this purpose. The gradient and Hessian are computed by analytic differentiation of $s(t)$.

\subsection{Object Tracking Framework}
\label{sec:object_tracking}
We first present the object tracking framework based on our continuous learning formulation introduced in section~\ref{sec:learning_formulation}. We employ multi-resolution feature maps $x_j$ extracted from a pre-trained deep network.\footnote{We use imagenet-vgg-m-2048, available at: \url{http://www.vlfeat.org/matconvnet/}.}
Similar to DCF based trackers \cite{DanelljanICCV2015,DanelljanCVPR14,Henriques14}, we extract a single training sample $x_j$ in each frame. The sample is extracted from an image region centered at the target location and the region size is set to $5^2$ times the area of the target box. Its corresponding importance weight is set to $\alpha_j = \frac{\alpha_{j-1}}{1 - \lambda}$ using a learning rate parameter $\lambda = 0.0075$. The weights are then normalized such that $\sum_j \alpha_j = 1$. We store a maximum of $m = 400$ samples by replacing the sample with the smallest weight. The Fourier coefficients $\hat{w}$ of the penalty function $w$ are computed as described in \cite{DanelljanICCV2015}. To detect the target, we perform a multi-scale search strategy \cite{DanelljanICCV2015,Li2014} with $5$ scales and a relative scale factor $1.02$. The extracted confidences are maximized using the grid search followed by five Newton iterations, as described in section~\ref{sec:localization}.

The training of our continuous convolution filter $f$ is performed by iteratively solving the normal equations \eqref{eq:normal_eq}. The work of \cite{DanelljanICCV2015} employed the Gauss-Seidel method for this purpose. However, this approach suffers from a quadratic complexity $\ordo(D^2)$ in the number of feature channels $D$. Instead, we employ the Conjugate Gradient (CG) \cite{NumericalOptimization} method due to its computational efficiency. Our numerical optimization scales linearly $\ordo(D)$ and is therefore especially suitable for high-dimensional deep features. In the first frame, we use $100$ iterations to find an initial estimate of the filter coefficients $\vecft{f}$. Subsequently, $5$ iterations per frame are sufficient by initializing CG with the current filter.\footref{supp_det}

\subsection{Feature Point Tracking Framework}
\label{sec:feature_tracking}

Here, we describe the feature point tracking framework based on our learning formulation.
For computational efficiency, we assume a single-channel feature map ($D=1$), e.g. a grayscale image, and a constant penalty function $w(t) = \beta$. Under these assumptions, the normal equations \eqref{eq:normal_eq} form a diagonal system of equations. The filter coefficients are directly obtained as,
\begin{equation}
	\label{eq:feature_filter}
	\hat{f}[k] = \frac{\sum_{j=1}^{M} \alpha_j \overline{X_j[k] \hat{b}[k]} \hat{y}_j[k]}{\sum_{j=1}^{M} \alpha_j \big|X_j[k] \hat{b}[k]\big|^2 + \beta^2} \;, \quad k = -K, \ldots, K \,.
\end{equation}
Here, we have dropped the feature dimension index for the sake of clarity. In this case (single feature channel and constant penalty function), the training equation \eqref{eq:feature_filter} resembles the original MOSSE filter \cite{MOSSE2010}. However, our continuous formulation has several advantages compared to the original MOSSE. Firstly, our formulation employs an implicit interpolation model, given by $\hat{b}$. Secondly, each sample is labeled by a continuous-domain confidence $y_j$, that enables sub-pixel information to be incorporated in the learning. Thirdly, our convolution operator outputs continuous confidence functions, allowing accurate sub-pixel localization of the feature point. In our experiments, we show that the advantages of our continuous formulation are crucial for accurate feature point tracking.

%% file: experiments.tex
\section{Experiments}

We validate our learning framework for two applications: tracking of objects and feature points. For object tracking, we perform comprehensive experiments on three datasets: OTB-2015 \cite{OTB2015}, Temple-Color \cite{TempleColor}, and VOT2015 \cite{VOT2015}. For feature point tracking, we perform extensive experiments on the MPI Sintel dataset \cite{butler2012naturalistic}.

\begin{table}[!t]
	\centering
	\caption{A baseline comparison when using different combinations of convolutional layers in our object tracking framework. We report the mean OP ($\%$) and AUC ($\%$) on the OTB-2015 dataset. The best results are obtained when combining all three layers in our framework. The results clearly show the importance of multi-resolution deep feature maps for improved object tracking performance.}
	\resizebox{0.9\textwidth}{!}{%
		\input{tables/baseline.tex}
	}
	\label{tab:baseline}%
\end{table}

\subsection{Baseline Comparison}
\label{sec:baseline}

We first evaluate the impact of fusing multiple convolutional layers from the deep network in our object tracking framework.
Table~\ref{tab:baseline} shows the tracking results, in mean overlap precision (OP) and area-under-the-curve (AUC), on the OTB-2015 dataset. OP is defined as the percentage of frames in a video where the intersection-over-union overlap exceeds a threshold of $0.5$. AUC is computed from the success plot, where the mean OP over all videos is plotted over the range of thresholds $[0, 1]$. For details about the OTB protocol, we refer to \cite{Wu13}.

In our experiments, we investigate the impact of the input RGB image layer (layer 0), the first convolutional layer (layer 1) and the last convolutional layer (layer 5). No significant gain in performance was observed when adding intermediate layers. The shallow layer (layer 1) alone provides superior performance compared to using only the deep convolutional layer (layer 5). Fusing the shallow and deep layers provides a large improvement. The best results are obtained when combining all three convolutional layers in our learning framework. We employ this three-layer combination for all further object tracking experiments.

We also compare our continuous formulation with the discrete DCF formulation by performing explicit resampling of the feature layers to a common resolution. For a fair comparison, all shared parameters are left unchanged. The layers (0, 1 and 5) are resampled with bicubic interpolation such that the data size of the training samples is preserved. On OTB-2015, the discrete DCF with resampling obtains an AUC score of $47.7\%$, compared to $68.2\%$ for our continuous formulation. This dramatic reduction in performance is largely attributed to the reduced resolution in layer 1. To mitigate this effect, we also compare with only resampling layers 0 and 5 to the resolution of layer 1. This improves the result of the discrete DCF to $60.8\%$ in AUC, but at the cost of a 5-fold increase in data size. Our continuous formulation still outperforms the discrete DCF as it avoids artifacts introduced by explicit resampling.

\begin{table}[!t]
	\centering
	\caption{A Comparison with state-of-the-art methods on the OTB-2015 and Temple-Color datasets. We report the mean OP ($\%$) for the top 10 methods on each dataset.
		Our approach outperforms DeepSRDCF by $5.1\%$ and $5.0\%$ respectively.}%
	\resizebox{1.015\textwidth}{!}{%
		\input{tables/OTB_TPL.tex}
	}%
	\label{tab:OTB_TPL}%
	\vspace{-3mm}
\end{table}
\begin{figure}[!t]
	\centering%
	\newcommand{\wid}{0.45\textwidth}
	\subfloat[OTB-2015\label{fig:sota_otb}]{\includegraphics[width = \wid]{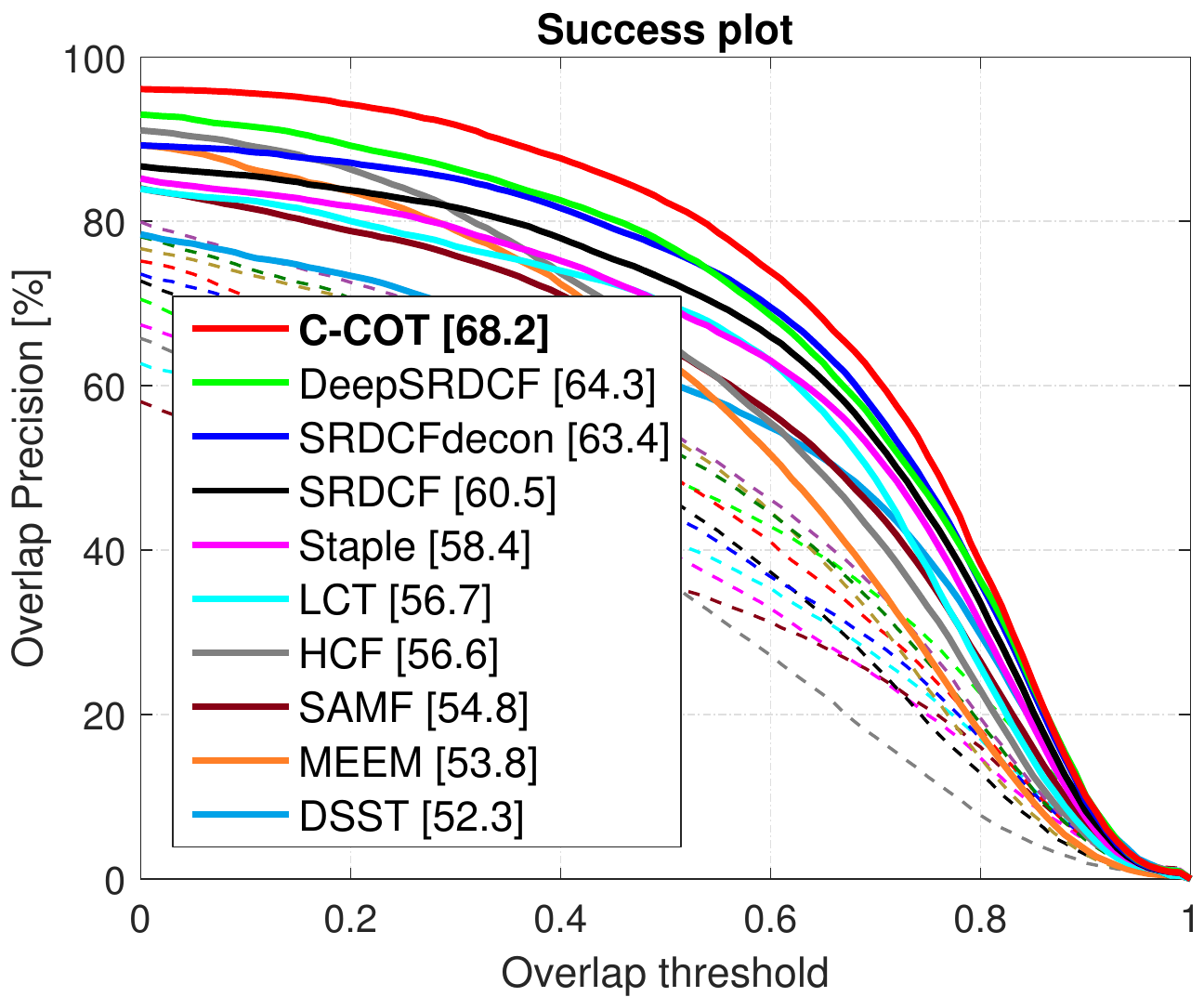}}\hspace{2mm}
	\subfloat[Temple-Color\label{fig:sota_tpl}]{\includegraphics[width = \wid]{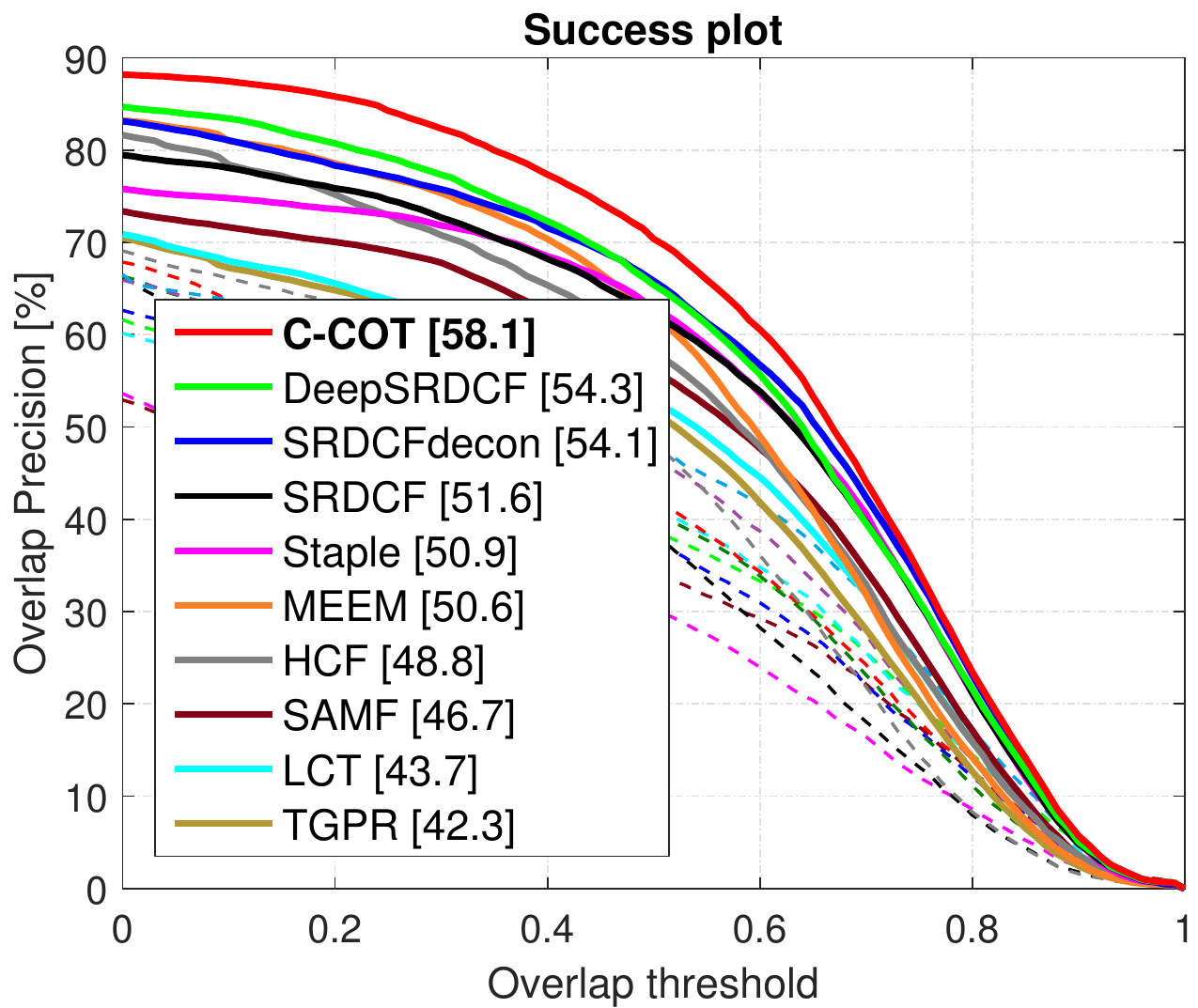}}%
	\caption{Success plots showing a comparison with state-of-the-art on the OTB-2015 (a) and Temple-Color (b) datasets. Only the top 10 trackers are shown for clarity. Our approach improves the state-of-the-art by a significant margin on both these datasets.}%
	\label{fig:OTB_tpl}
\end{figure}

\subsection{OTB-2015 Dataset}

We validate our Continuous Convolution Operator Tracker (C-COT) in a comprehensive comparison with 20 state-of-the-art methods: ASLA \cite{Jia12d}, TLD \cite{Mikolajczyk10d}, Struck \cite{Torr11b}, LSHT \cite{Shengfeng13b}, EDFT \cite{Felsberg13c}, DFT \cite{Laura12d}, CFLB \cite{GaloogahiCVPR2015}, ACT \cite{DanelljanCVPR14}, TGPR \cite{TGPR2014}, KCF \cite{Henriques14}, DSST \cite{DanelljanBMVC14}, SAMF \cite{Li2014}, MEEM \cite{MEEM2014}, DAT \cite{possegger15a}, LCT \cite{LTC_CVPR15}, HCF \cite{HCF_ICCV15}, Staple \cite{Staple} and SRDCF \cite{DanelljanICCV2015}. We also compare with SRDCFdecon, which integrates the adaptive decontamination of the training set \cite{DanelljanCVPR2016a} in SRDCF, and DeepSRDCF \cite{DanelljanVOT2015} employing activations from the first convolutional layer.

\parsection{State-of-the-art Comparison}
Table~\ref{tab:OTB_TPL} (first row) shows a comparison with state-of-the-art methods on the OTB-2015 dataset.\footlabel{results}{Detailed results are provided in the supplementary material.} The results are reported as mean OP over all the 100 videos. The HCF tracker, based on hierarchical convolutional features, obtains a mean OP of $65.5\%$. The DeepSRDCF employs the first convolutional layer, similar to our baseline ``Layer 1'' in table~\ref{tab:baseline}, and obtains a mean OP of $77.3\%$. Our approach achieves the best results with a mean OP of $82.4\%$, significantly outperforming DeepSRDCF by $5.1\%$.
 
Figure~\ref{fig:sota_otb} shows the success plot on the OTB-2015 dataset. We report the AUC score for each tracker in the legend. The DCF-based trackers HCF and Staple obtain AUC scores of $56.6\%$ and $58.4\%$ respectively. Among the compared methods, the SRDCF and its variants SRDCFdecon and DeepSRDCF provide the best results, all obtaining AUC scores above $60\%$. Overall, our tracker achieves the best results, outperforming the second best method by $3.9\%$.

\begin{figure}[!t]
	\centering
	\newcommand{\wid}{0.45\textwidth}
	\includegraphics[width = \wid]{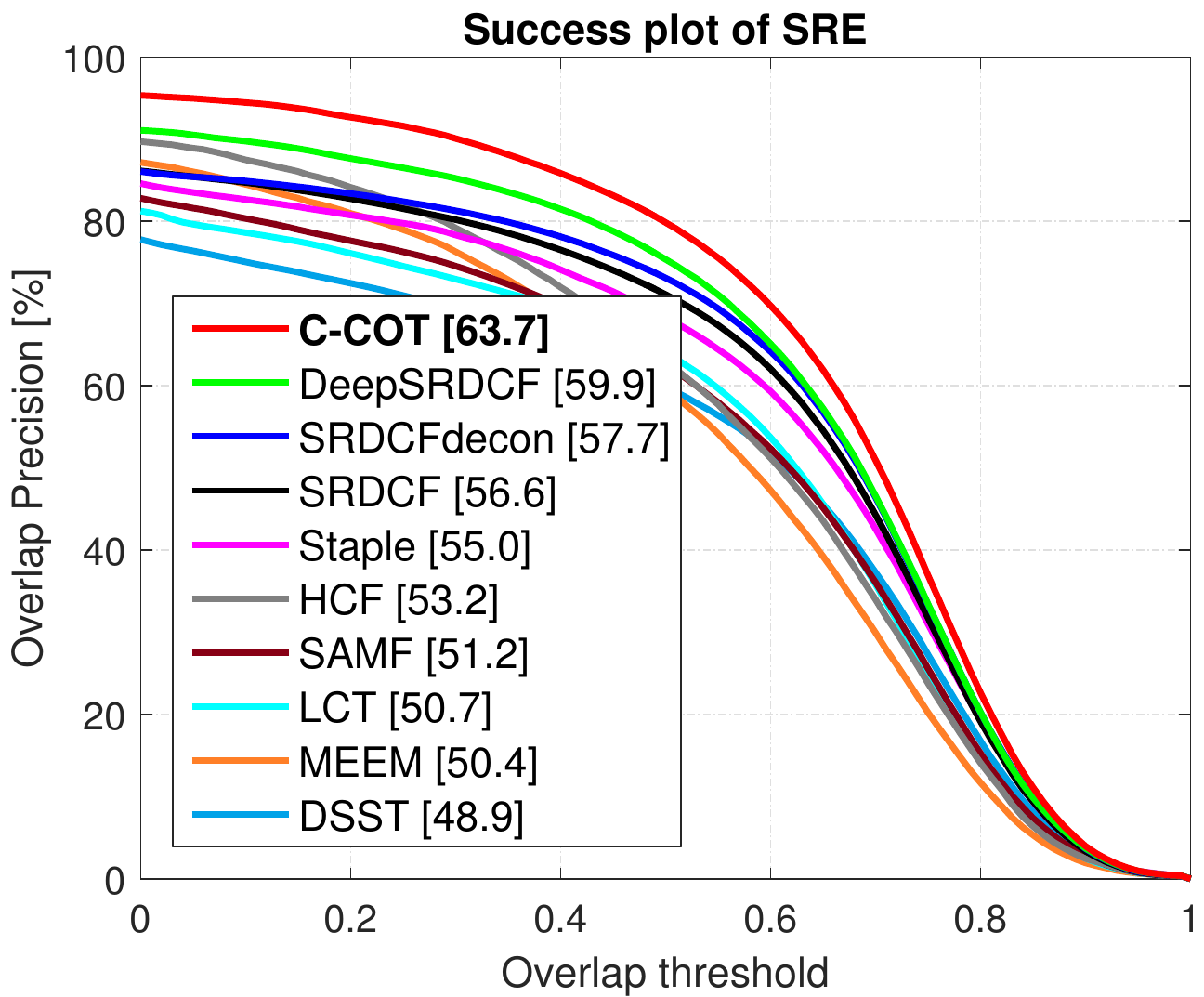}\hspace{2mm}
	\includegraphics[width = \wid]{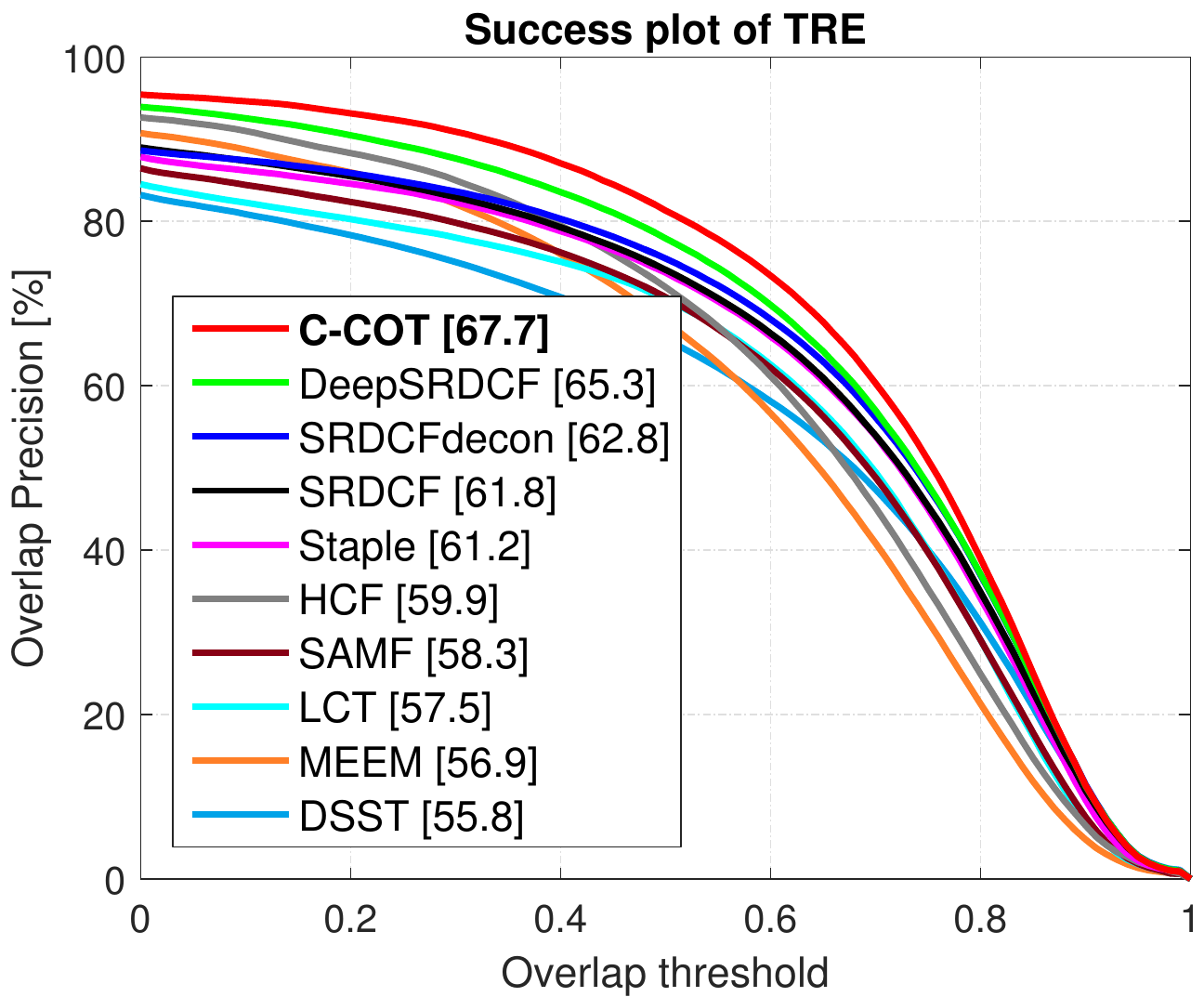}
	\caption{An evaluation of the spatial (\textit{left}) and temporal (\textit{right}) robustness to initializations on the OTB-2015 dataset. We compare the top 10 trackers. Our approach demonstrates superior robustness compared to state-of-the-art methods.}%
	\label{fig:sota_sretre}
\end{figure}

\parsection{Robustness to Initialization}
We evaluate the robustness to initializations using the protocol provided by \cite{OTB2015}. Each tracker is evaluated using two different initialization strategies: spatial robustness (SRE) and temporal robustness (TRE). The SRE criteria initializes the tracker with perturbed boxes, while the TRE criteria starts the tracker at 20 frames. Figure~\ref{fig:sota_sretre} provides the SRE and TRE success plots. Our approach obtains consistent improvements in both cases.

\subsection{Temple-Color Dataset}
Here, we evaluate our approach on the Temple-Color dataset \cite{TempleColor} containing 128 videos. The second row of table~\ref{tab:OTB_TPL} shows a comparison with state-of-the-art methods. The DeepSRDCF tracker provides a mean OP score of $65.4\%$. MEEM and SRDCFdecon obtain mean OP scores of $62.2\%$ and $65.8\%$ respectively. Different from these methods, our C-COT does not explicitly manage the training set to counter occlusions and drift. Our approach still improves the start-of-the-art by a significant margin, achieving a mean OP score of $70.4\%$. A further gain in performance is expected by incorporating the unified learning framework \cite{DanelljanCVPR2016a} to handle corrupted training samples. In the success plot in Figure~\ref{fig:sota_tpl}, our method obtains an absolute gain of $3.8\%$ in AUC compared to the previous best method.

\subsection{VOT2015 Dataset}
The VOT2015 dataset \cite{VOT2015} consists of 60 challenging videos compiled from a set of more than 300 videos. Here, the performance is measured both in terms of accuracy (overlap with the ground-truth) and robustness (failure rate). In VOT2015, a tracker is restarted in the case of a failure. We refer to \cite{VOT2015} for details. Table~\ref{tab:VOT} shows the comparison of our approach with the top 10 participants in the challenge according to the VOT2016 rules \cite{VOT2016}. Among the compared methods, RAJSSC achieves favorable results in terms of accuracy, at the cost of a higher failure rate. EBT achieves the best robustness among the compared methods. Our approach improves the robustness with a $20\%$ reduction in failure rate, without any significant degradation in accuracy.

\begin{table}[!t]
	\centering
	\caption{Comparison with state-of-the-art methods on the VOT2015 dataset. The results are presented in terms of robustness and accuracy. Our approach provides improved robustness with a significant reduction in failure rate.}%
	\resizebox{1.015\textwidth}{!}{%
		\input{tables/VOT.tex}
	}
	\label{tab:VOT}
\end{table}

\subsection{Feature Point Tracking}
We validate our approach for robust and accurate feature point tracking. Here, the task is to track distinctive local image regions. We perform experiments on the MPI Sintel dataset \cite{butler2012naturalistic}, based on the 3D-animated movie ``Sintel''. The dataset consists of 23 sequences, featuring naturalistic and dynamic scenes with realistic lighting and camera motion blur. The ground-truth dense optical flow and occlusion maps are available for each frame.
Evaluation is performed by selecting approximately 2000 feature points in the first frame of each sequence. We use the Good Features to Track (GFTT) \cite{shi1994gftt} feature selector, but discard points at motion boundaries due to their ambiguous motion. The ground-truth tracks are then generated by integrating flow vectors over the sequence. The flow vectors are obtained by a bilinear interpolation of the dense ground-truth flow. We terminate the ground-truth tracks using the provided occlusion maps. 

We compare our approach to MOSSE \cite{MOSSE2010} and KLT \cite{LukasKanade81,tomasi1991KLTtracker}. The OpenCV implementation of KLT, used in our experiments, employs a pyramidal search \cite{bouguet2000klt} to accommodate for large translations. For a fair comparison, we adopt a similar pyramid approach for our method and MOSSE, by learning an independent filter for each pyramid level. Further, we use the window size of $31 \times 31$ pixels and $3$ pyramid levels for all methods. For both our method and MOSSE we use a learning rate of $\lambda = 0.1$ and set the regularization parameter to $\beta = 10^{-4}$. For the KLT we use the default settings in OpenCV. Unlike ours and the MOSSE tracker, the KLT tracks feature points frame-to-frame without memorizing earlier appearances. In addition to our standard tracker, we also evaluate a frame-to-frame version (Ours-FF) of our method by setting the learning rate to $\lambda = 1$. 

\begin{figure}[!t]
	\centering%
	\newcommand{\wid}{2.7cm}%
	\includegraphics*[trim=0 0 0 0,height = \wid]{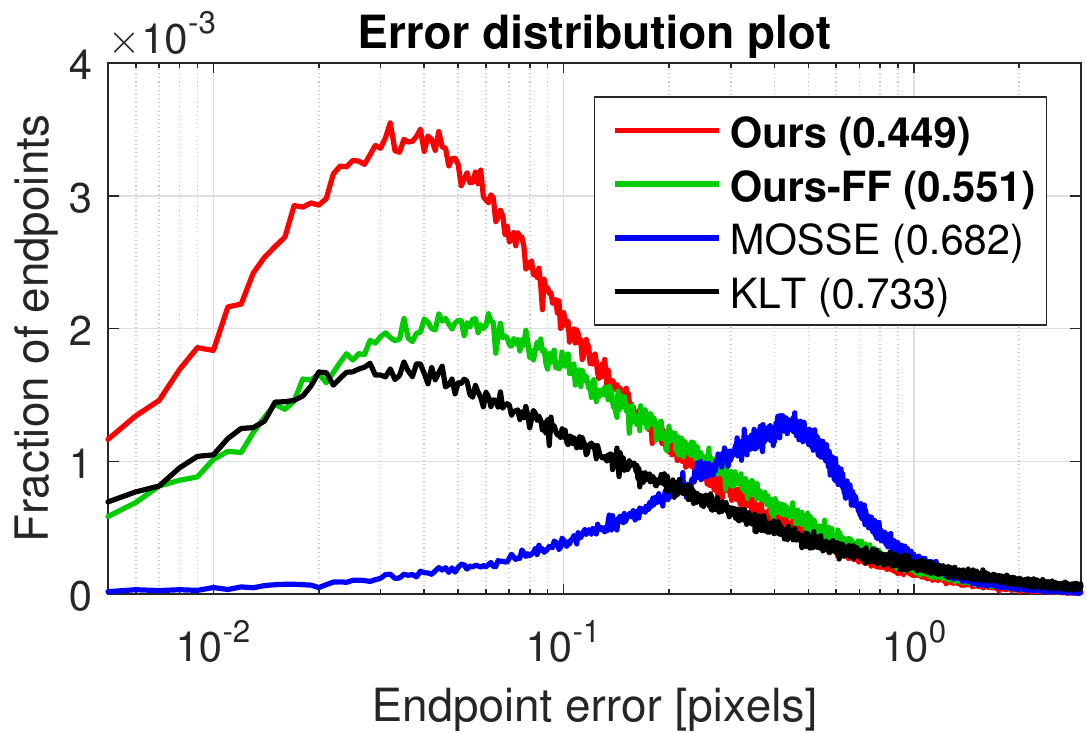}\hspace{1mm}%
	\includegraphics*[trim=0 0 0 0,height = \wid]{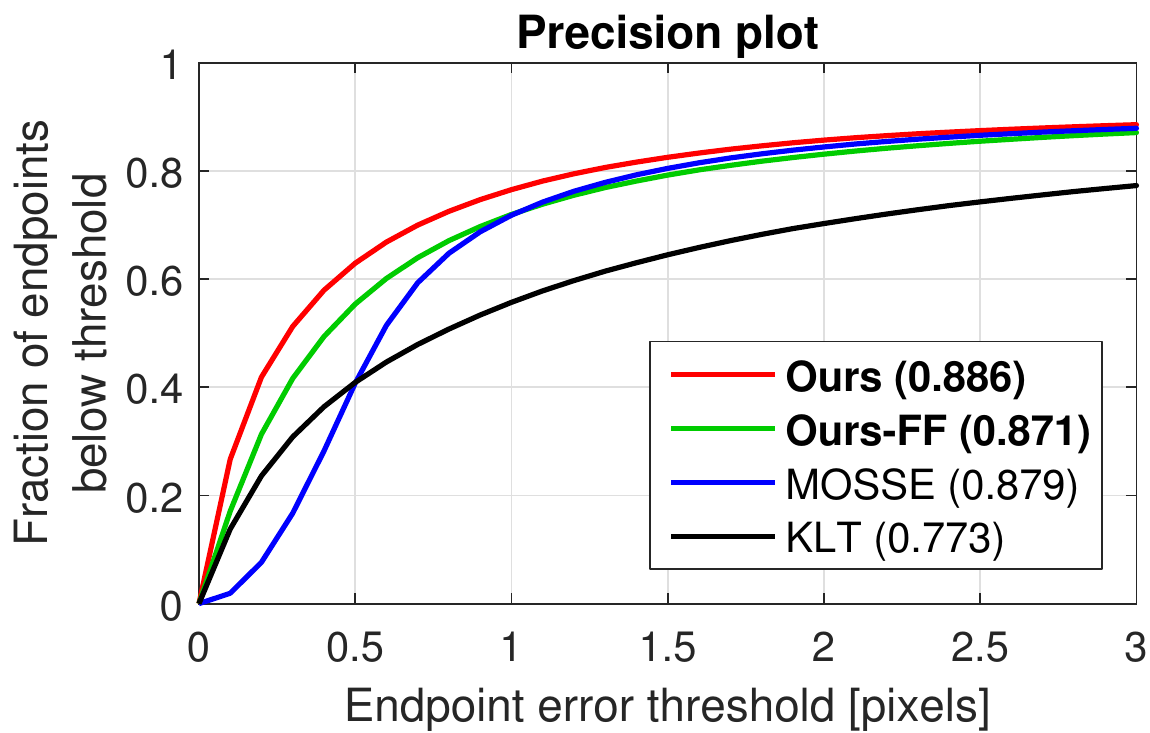}\hspace{1mm}%
	\includegraphics*[trim=170 0 150 0,height=\wid]{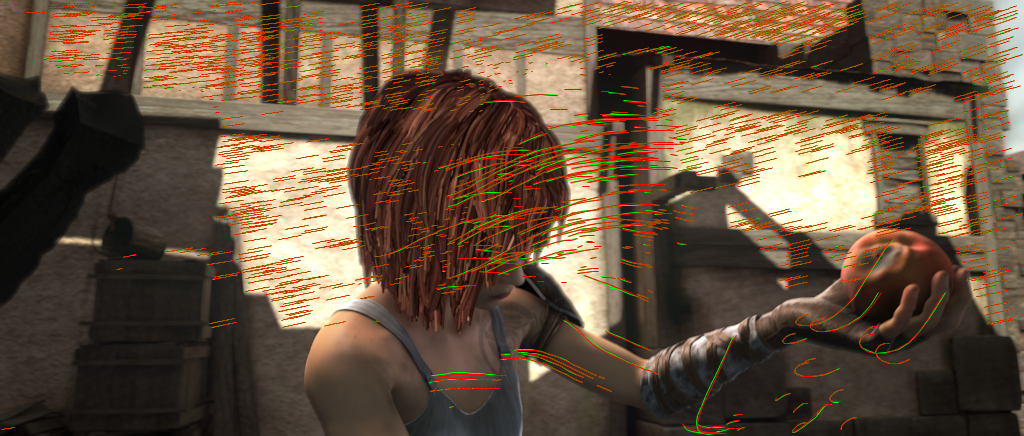}%
	\caption{Feature point tracking results on the MPI Sintel dataset. We report the endpoint error (EPE) distribution (\textit{left}) and precision plot (\textit{center}) over all sequences and points. In the legends, we display the average inlier EPE and the inlier ratio for the error distribution and precision plot respectively. Our approach provides consistent improvements, both in terms of accuracy and robustness, compared to existing methods. The example frame (\textit{right}) from the Sintel dataset visualizes inlier trajectories obtained by our approach (red) along with the ground-truth (green).}%
	\label{fig:pt_tracking}
\end{figure}

For quantitative comparisons, we use the endpoint error (EPE), defined as the Euclidian distance between the tracked point and its corresponding ground-truth location. Tracked points with an EPE smaller than 3 pixels are regarded as inliers. Figure~\ref{fig:pt_tracking} (left) shows the distribution of EPE computed over all sequences and tracked points. We also report the average inlier EPE for each method in the legend. Our approach achieves superior accuracy, with an inlier error of $0.449$ pixels. We also provide the precision plot (Figure~\ref{fig:pt_tracking}, center), where the fraction of points with an EPE smaller than a threshold is plotted. The legend shows the inlier ratio for each method. Our tracker achieves superior robustness in comparison to the KLT, with an inlier ratio of $0.886$. Compared to MOSSE, our method obtains significantly improved precision at sub-pixel thresholds ($<1$ pixel). This clearly demonstrates that our continuous formulation enables accurate sub-pixel feature point tracking, while being robust. Unlike the frame-to-frame KLT, our method provides a principled procedure for updating the tracking model, while memorizing old samples. The experiments show that already our frame-to-frame variant (Ours-FF) provides a spectacular improvement compared to the KLT. Hence, our gained performance is due to both the model update \emph{and} the proposed continuous formulation. On a desktop machine, our Matlab code achieves real-time tracking of 300 points at a single scale, utilizing only a single CPU.

%% file: tables/baseline.tex
\begin{tabular}{l@{~}|@{~}c@{~~}c@{~~}c@{~}|@{~}c@{~~}c@{~~}c@{~}|@{~}c}
\toprule
&Layer 0&Layer 1&Layer 5&Layers 0, 1&Layers 0, 5&Layers 1, 5&Layers 0, 1, 5\\\midrule
Mean OP&58.8&78.0&60.0&77.8&70.7&\textit{\textcolor{blue}{81.8}}&\textbf{\textcolor{red}{82.4}}\\
AUC&49.9&65.8&51.1&65.7&59.0&\textit{\textcolor{blue}{67.8}}&\textbf{\textcolor{red}{68.2}}\\\bottomrule
\end{tabular}

%% file: tables/OTB_TPL.tex
\begin{tabular}{l@{~}c@{~~}c@{~~}c@{~~}c@{~~}c@{~~}c@{~~}c@{~~}c@{~~}c@{~~}c@{~~}c}
\toprule
&DSST&SAMF&TGPR&MEEM&LCT&HCF&Staple&SRDCF&SRDCFdecon&DeepSRDCF&\textbf{C-COT}\\\midrule
OTB-2015&60.6&64.7&54.0&63.4&70.1&65.5&69.9&72.9&76.7&\textit{\textcolor{blue}{77.3}}&\textbf{\textcolor{red}{82.4}}\\
Temple-Color&47.5&56.1&51.6&62.2&52.8&58.2&63.0&62.2&\textit{\textcolor{blue}{65.8}}&65.4&\textbf{\textcolor{red}{70.4}}\\\bottomrule
\end{tabular}

%% file: tables/VOT.tex
\begin{tabular}{l@{~}c@{~~}c@{~~}c@{~~}c@{~~}c@{~~}c@{~~}c@{~~}c@{~~}c@{~~}c@{~~}c}
\toprule
&S3Tracker&RAJSSC&Struck&NSAMF&SC-EBT&sPST&LDP&SRDCF&EBT&DeepSRDCF&\textbf{C-COT}\\\midrule
Robustness&1.77&1.63&1.26&1.29&1.86&1.48&1.84&1.24&1.02&1.05&0.82\\
Accuracy&0.52&0.57&0.47&0.53&0.55&0.55&0.51&0.56&0.47&0.56&0.54\\\bottomrule
\end{tabular}

%% file: conclusions.tex
\section{Conclusions} 
We propose a generic framework for learning discriminative convolution operators in the continuous spatial domain. We validate our framework for two problems: object tracking and feature point tracking. Our formulation enables the integration of multi-resolution feature maps. In addition, our approach is capable of accurate sub-pixel localization. Experiments on three object tracking benchmarks demonstrate that our approach achieves superior performance compared to the state-of-the-art. Further, our method obtains substantially improved accuracy and robustness for real-time feature point tracking.

Note that, in this work, we do not use any video data to learn an application specific deep feature representation. This is expected to further improve the performance of our object tracking framework. Another research direction is to incorporate motion-based deep features into our framework, similar to \cite{GladhICPR2016}. 

\noindent\textbf{Acknowledgments}:
This work has been supported by SSF (CUAS), VR (EMC${}^2$), CENTAURO, the Wallenberg Autonomous Systems Program, NSC and Nvidia.